\documentclass[letterpaper, 10 pt, conference]{ieeeconf}  

\IEEEoverridecommandlockouts                              

\usepackage{multicol}
\usepackage{multirow}
\usepackage[bookmarks=true]{hyperref}
\usepackage{amssymb}
\usepackage{tabularx}
\usepackage{amsmath}
\usepackage{caption}
\usepackage{algorithm}

\usepackage{algpseudocode}
\usepackage{subcaption}
\usepackage{float}
\usepackage[backend=bibtex,style=ieee]{biblatex}
\bibliography{references_v1.bib}

\title{\LARGE \bf
\textbf{Learning to Recover from Plan Execution Errors during Robot Manipulation: A Neuro-symbolic Approach}
}
\author{
Namasivayam K$^{*}$$^{1}$,
Arnav Tuli$^{*}$$^{2}$, 
Vishal Bindal$^{*}$$^{2}$, 
Himanshu Singh$^{2}$, 
Parag Singla$^{1}$ and Rohan Paul$^{1}$\\
\small $^{1}$Affiliated with IIT Delhi. 
\small $^{2}$Work done when at IIT Delhi.
$^{*}$ denotes equal contribution.
}

\usepackage{contour}
 \usepackage{array,multirow,graphicx}
 \usepackage{float}
 \newcolumntype{P}[1]{>{\centering\arraybackslash}m{#1}}
\usepackage{dsfont}
\contourlength{0.8pt}

\DeclareMathOperator*{\argmin}{arg\,min}

\begin{document}

\maketitle
\thispagestyle{empty}
\pagestyle{empty}
\begin{abstract}
Automatically detecting and recovering from failures is an important but challenging problem for autonomous robots. Most of the recent work on learning to plan from demonstrations lacks the ability to detect and recover from errors in the absence of an explicit state representation and/or a (sub-) goal check function. We propose an approach (blending learning with symbolic search) for automated error discovery and recovery, without needing annotated data of failures. Central to our approach is a neuro-symbolic state representation, in the form of dense scene graph, structured based on the objects present within the environment. This enables efficient learning of the transition function and a discriminator that not only identifies failures but also localizes them facilitating fast re-planning via computation of heuristic distance function. We also present an anytime version of our algorithm, where instead of recovering to the last correct state, we search for a sub-goal in the original plan minimizing the total distance to the goal given a re-planning budget. Experiments on a physics simulator with a variety of simulated failures show the effectiveness of our approach compared to existing baselines, both in terms of efficiency as well as accuracy of our recovery mechanism.
\end{abstract}
\section{Introduction}\label{sec:intro}
An essential aspect of robot plan execution is \emph{recovering} from errors. While plans may be perfect in simulation, they are often marred by imperfections in the real world caused by sensor noise, motor failures, unexpected collisions or external disturbances. This results in deviations from the original path of planned execution, which the robot needs to recover from.
Once an error is detected, an efficient plan repair is necessary to return to a nominal state in the original plan, accommodating progress made before the error. 
%

Error recovery in classical task planning~\cite{fox2006plan,bercher2014plan,saetti2022optimising} detect errors by explicit symbolic reasoning, such as checking the precondition of the next action in the plan, and recover through replanning to the original goal or last correct state.
%
%
Alternate approaches for planning under uncertainty use reinforcement learning~\cite{thananjeyan2021recovery,vatsefficient,li2021reactive} to learn a reactive policy for the robot to recover from a deviant state to a nominal state. However, learning such a policy requires extensive offline exploration of error states, which is challenging in complex manipulation domains. 

Recently, \emph{learning-to-plan} methods 
have been proposed that predict a plan - essentially a sequence of symbolic actions - to reach a goal specified through language instruction \cite{Kalithasan2022LearningNP, wang2023progport, zhu2021hierarchical, xu2019regression} or logical expression\cite{mao2022pdsketch}. They are trained in a supervised manner via a dataset comprising demonstrations of goal reaching plans. 
%
A key limitation in these approaches is the absence of a feedback mechanism to assess goal attainment, resulting in their inability to detect execution errors. Even when the model has such an ability detect whether the goal has been attained, they require additional annotation data to learn such a goal-check function~\cite{mao2022pdsketch}. Further, there is no way to monitor the correctness of the intermediate states during the execution. In such mechanisms, the error recovery can be cast as re-planning \emph{directly} to the goal, albeit \emph{disregarding} prior planning effort.

\begin{figure}[t!]
    \begin{subfigure}{1.0\hsize}
         \centering    
         \includegraphics[width =\textwidth, height = 4.2cm]{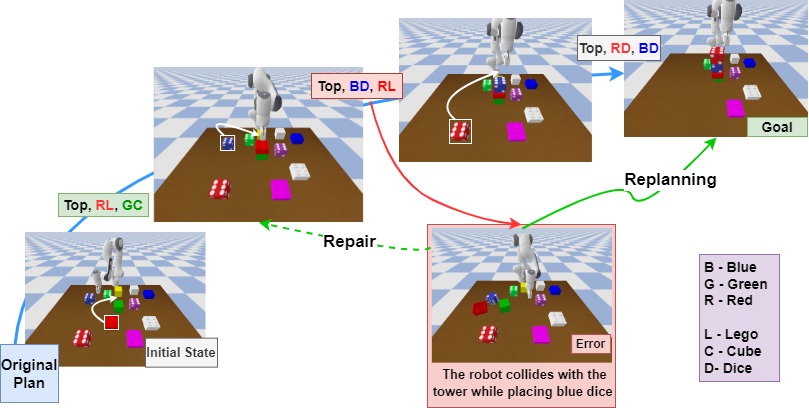}
    \end{subfigure}
    \caption{
            \textbf{Error Recovery Problem.}  
        The diagram illustrates an error resulting from the robot colliding with a partially constructed tower. The states along the blue line represent the expected progression in the plan execution devoid of errors. However, the collision results in a significant deviation from this expected path, requiring error recovery planning.
    }
    \vspace{-0.15in}
    \label{fig:nsrm}
\end{figure}

Our goal is to address the challenge of monitoring and recovering from errors during the execution of multi-step task plans generated by learning-to-plan methods. We aim to learn to detect and recover from errors solely using demonstrations of goal-reaching plans without any need for annotated data of failures.
Figure 1 illustrates a situation where the robot is assigned the task of constructing a tower by stacking blocks on top of each other. During its execution, the robot collides with a portion of the tower, causing the entire structure to collapse. Not only the robot has to be able to detect that error has occurred, it also needs to recover from the error by constructing a multi-step plan to re-stack the blocks from the current (erroneous) state. 

In general, designing such an error recovery mechanism opens up two key questions: (a) \emph{``Can we learn a state discrimination function capable of identifying and localising differences between states in a self-supervised way without the need for manually annotated data?"} 
(b) \emph{``Can we leverage localised information about the failure to efficiently reach the original goal while minimising the replanning time and re-use a part of the original plan as appropriate?"}. 

In response, we propose an approach which answers both the above questions in affirmative. 
%
We represent the world as a scene graph over neural object representations. 
Our key insight is to learn (i) a scene-graph transition model predicting the next scene graph for an input action; providing a foresight into how the scene will look as the plan is executed, and (ii) \emph{neural discriminators} to distinguish between two states as well as object representations. 
The discriminator trained in a \emph{self-supervised} way serves a dual role: it identifies the objects that are responsible for the failure and serves as a heuristic between states by discriminating object representations. 
Once an error is detected, the recovery plan is generated through a directed plan search toward a sub-goal in the original plan that is nearest to the error state, while being faithful to the progress achieved prior to the error. The search for the recovery plan is (i) \emph{neuro-symbolic} with explicit symbolic search operating over neural action models and (ii) \emph{discrepancy}-aware using context of erroneous object states during plan search.  

The approach is evaluated on a data set of simulated robot execution errors such as grasp failure during transport, adversarial/cooperative interventions in the scene and non-determinism in action outcomes due to motion planning errors. 
Experiments demonstrate significantly better recovery in multi-step plans marred by sequential errors 
compared to baselines consisting of an RL-based reactive error recovery mechanism and a fully re-planning based strategy inspired by Mao et al.~\cite{mao2022pdsketch}, both in terms of recovery rate, and the time taken to reach the sub-goal for recovery.   
\section{Related Works} \label{sec:related}
Robotic systems deployed with symbolic task planners~\cite{knepper2013ikeabot,hudson2012end} monitor plan execution by repeatedly checking for errors between the intended and the actual states encountered. Formalisms such as ~\cite{bercher2014plan,saetti2022optimising,fox2006plan} synthesize a plan closest to the original or return back to the last correct state. Others, maintain a library of hand-designed recovery behaviours \cite{tellex2014asking,huang2019neural} in anticipation of failures switching to recovery behaviours when errors occur. Such strategies inherit the same brittleness as pure symbolic planning due imperfections in extracting state knowledge from sensor data. 

Alternative RL-based approaches learn reactive neural policies that prescribe an action for a given state that the robot may encounter~\cite{rana2023residual,kumar2023graph,ebert2018visual,ryu2022confidence}. Notable successes include learning manipulation skills. In this paradigm, error recovery is \emph{implicit} in the learned policy, if during learning, the agent has explored the state space \emph{well enough} so as to  generalize to any state that the robot may encounter online. Such generalization is difficult in complex manipulation domains and hence RL-based reactive planners are used for local error recovery or applied in simpler state spaces. 
%
%
Whereas such approaches attain \emph{myopic} short horizon recovery, our work addresses plan repair over a longer horizon inherent in complex manipulation task (e.g., recovery from a fallen stack of blocks). 

Recent \emph{learning-to-plan} methods fuse neural representations with symbolic reasoning for generalized long-horizon planning~\cite{mao2022pdsketch,Kalithasan2022LearningNP,zhu2021hierarchical,xu2019regression,shridhar2022cliport,Mao2019TheNC,sung2023learning}. 
Such planners allow data-driven learning of both spatial and action representations that can be composed for goal-directed reasoning. 
Error recovery in these models is less studied and largely accomplished by full re-planning to the goal. This paper presents a formalism for error recovery in such models. 
Our work is also closely related to~\cite{sung2023learning}, who explore plan recovery in the context of task and motion planning problems that additionally consider planning in metric space. Their approach predicts the cause of plan failure and performs back jumping to construct a recovery plan. However, this approach uses supervised learning to predict the cause for a plan failure using supervised learning with a data set of failed plans. In contrast, our work ameliorates the need for an explicit corpus of failures. 

%

\section{Background and Problem Statement}\label{sec:problem}
We consider a robot in a tabletop setting that observes the environment through a depth sensor. The robot is capable of positioning its end-effector at a pose, and grasping/releasing the object at a given pose.
The robot is provided high-level goals such as stacking objects of a certain colour, moving objects to a certain region, etc. 
We assume that the robot possesses a high-level task planner $\mathcal{P}$ that operates on spatial abstractions such as $\mathrm{Top()}, \mathrm{Left()}$ and $\mathrm{Right()}$ and can synthesize a plan as a sequence of actions such as $\mathrm{Move()}$, $\mathrm{Place()}$ etc. 
Such representations lie at the core of learning-to-plan planning systems such as \cite{Kalithasan2022LearningNP,mao2022pdsketch, wang2023progport, xu2019regression}.  
Formally, given a high-level goal specification, $\mathcal{P}$ generates a plan $\Pi = (a_1, a_2, .., a_T)$, where $a_i$ is an action from $\mathcal{A}$ and $\mathcal{A}$ denotes the high-level actions. 
Further, we assume a data set $\mathbb{D}$ consisting of goal-reaching plans. 
%

We model the execution of the given plan as a goal-conditioned Markov Decision Process (MDP), denoted by $\big < \mathcal{S}, \mathcal{A}, \mathcal{T}^{env}, S_I, {g}, \mathcal{R} \big >$. 
We assume an object-centric state space: a state $S\in \mathcal{S}$ is characterized by the appearance (colour, material etc) and the metric location of the rigid objects in the scene. 
The action space comprises high-level actions with discrete parameters, such as $\mathtt{MoveTop(a, b)}$, intending to achieve the desired spatial relationship between the objects. 
The plan $\Pi$ defines an implicit goal $g$, representing the expected final state in an \emph{error-free} execution from the initial state $S_I$.
However, reaching the expected state in real robot executions are hindered by errors such as action execution issues (e.g., an object falling from the robot's gripper), collateral effects (e.g., disturbances causing objects to fall in a tower-like assembly), or actions by an external agent, such as a human (adversarial or cooperative).
The transition model $\mathcal{T}^{env}: \mathcal{S} \times \mathcal{A} \rightarrow \mathcal{S}$ unknown to the robot characterizes these erroneous transitions, determining the next state based on the current state and action during the actual robot execution.
The robot's objective is to execute the long-horizon plan $\Pi$ successfully, reaching the implicit goal state ${g}$ despite the presence of erroneous transitions $\mathcal{T}^{env}$ in the actual robotic executions.  
The robot is provided with a sparse reward for goal attainment. 

Our approach consists of two key components which are introduced next. The next section details learning architecture for detecting erroneous states by comparing ``imagined" future states with states encountered during execution. Section IV details the efficient synthesis of recovery plan once the error is detected. 

\section{Learning to Detect Error States}
Multi-step manipulation tasks involve complex inter-object interactions. We represent the state using a complete neural scene-graph, with nodes corresponding to objects in the scene.  Our error detection pipeline consists of the following predictors (detailed subsequently) trained using a dataset $\mathbb{D}$ comprising demonstrations of error-free plan executions (without explicit failure annotations). 
\begin{itemize}
\item \emph{Scene Graph Encoder}, $H_\psi$, that extracts a scene graph from the RGB-D image of the environment.

\item \emph{Scene graph prediction function:} 
$\mathcal{T}^{ideal}_\theta$ that estimates the \emph{intended} changes in the current scene graph when the robot takes an action. Successive application of this function predicts the future state sequence during nominal (error-free) plan execution.
\item The \emph{Scene graph discriminator function:}, $K_\phi$, that estimates the normalized degree of similarity between any two scene graphs. At every step of plan execution, the intended state on applying an action is imagined using the state transition function, and it is compared with the actual state reached using the discriminator; a significant discrepancy indicates an error. 
\end{itemize}

\subsection{Scene Graph Encoder $H_\psi$}
The state is represented by a neural scene graph extracted from visual data comprising an RGB-D image and object bounding boxes. Following~\cite{Mao2019TheNC, Kalithasan2022LearningNP, wang2023progport, zhu2021hierarchical}, the scene graph is factored in terms of the objects present in the scene. 
Visual features of each object are extracted using a pre-trained ResNet~\cite{He2016Resnet} based feature extractor~\cite{Mao2019TheNC, Kalithasan2022LearningNP}.
 The visual features for every object are concatenated with the bounding boxes (with depth) to include the metric (positional) information. The concatenated feature vector is passed through an MLP encoder ($E_N$), resulting in scene graph nodes that capture dense object representations
For each pair of objects, we form the \textit{edge-embedding} by simply concatenating the corresponding directed pair of \textit{node-embeddings}. The order of concatenation matters as the relations in real-world (like top, left and right) are not symmetric, and hence, there are two edges for every pair of objects, giving a total of $n \times (n - 1)$ edges, where $n$ is the number of objects. 

\subsection{Learning the Scene Graph Predictor $T^{ideal}_\theta$}
The scene-graph predictor predicts the changes in the scene graph when an action is applied. Actions typically have a local effect on the objects they are applied to. 
For instance, when a robot moves object A onto object B, only A's location and its relations with other objects change. Specifically, this involves modifying the node embeddings for A and the edge embeddings for edges of the form  (A, *) or (*, A). We assume actions in the action space $\mathcal{A}$ manipulate one object at a time. Consequently, the scene graph prediction at any given time step involves predicting a single node change. 

\begin{figure*}[h!]
    \begin{subfigure}{1.0\hsize}
         \centering    
         \includegraphics[width = \textwidth]{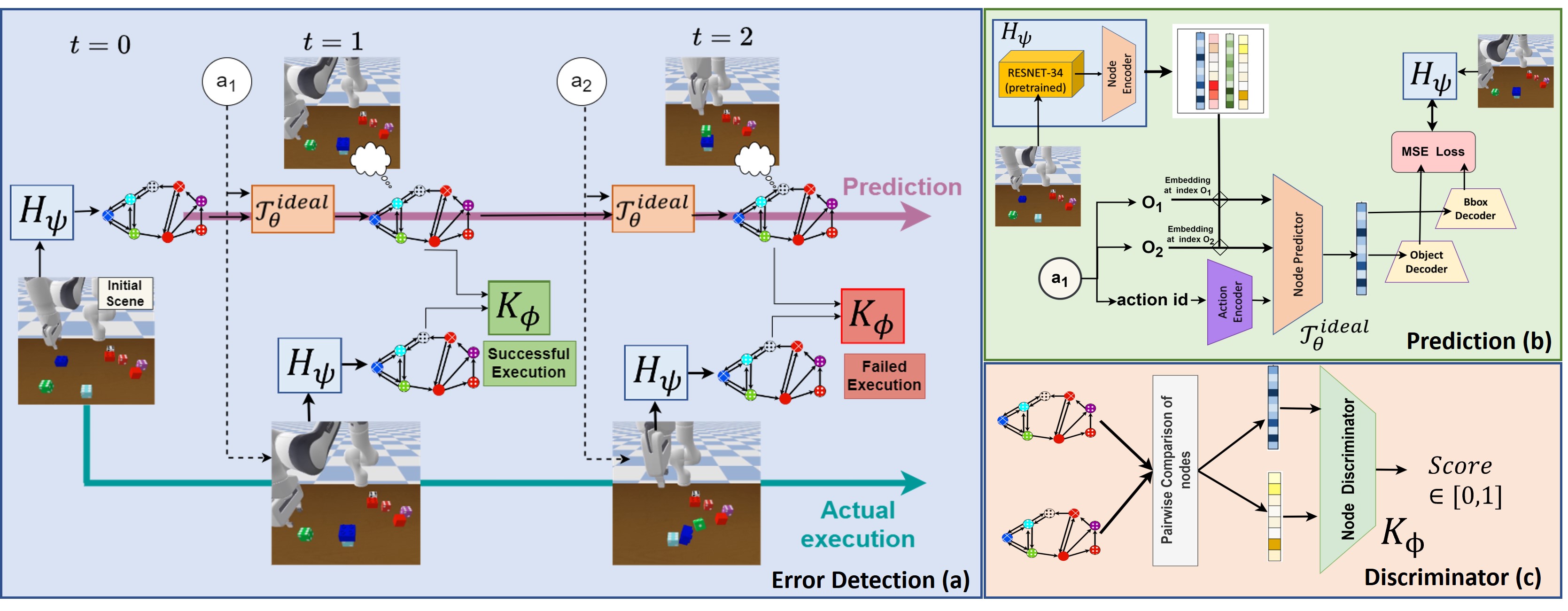}
    \end{subfigure}
    \caption{
            \textbf{(a) Model Overview:}
            The learnt scene graph predictor $\mathcal{T}^{ideal}_\theta$ is used to imagine the effect of an action execution, which is then compared with the actual scene graph to detect errors. Significant deviations are marked as error by Discriminator $K_\phi$, and the error recovery module will be triggered. \textbf{(b) Scene graph Predictor:} The architecture of scene-graph extractor and predictor (c) \textbf{Discriminator:} Architecture of the discriminator. 
        }
    \vspace{-0.1in}
    \label{fig:sgp_new}
\end{figure*}

Figure \ref{fig:sgp_new} describes the architecture of the scene-graph predictor. An action encoder transforms the one-hot action into a dense representation. These action features are then concatenated with the node embeddings of its arguments and passed through the node predictor, which predicts the new node embedding for the manipulated object. The predicted node embedding is fed into both the object and bounding box decoders to reconstruct the object embeddings and bounding box of the manipulated object. This reconstructed object embedding and bounding box are compared (using MSE-Loss) with the corresponding gold embeddings and bounding boxes extracted (using $H_\psi$) from the RGB-D image of the next state in the demonstration. All encoders and decoders are implemented as MLPs, with the weights $H_\psi$ and $\mathcal{T}^{ideal}_\theta$ jointly trained using the obtained loss.
%
%

The trained scene-graph predictor can iteratively imagine the scene graphs of future states encountered during the execution of plan $\Pi$. Let $\mathcal{E}(S_I, \Pi) = \mathcal{E}(\mathcal{T}^{ideal}(S_I, a_1), (a_2, a_3, \dots, a_T))$ be the final \emph{intended} state and $tr(\Pi)= (S_0 =S_I, S_1, \ldots, S_T=S_G)$ denote the \emph{nominal} error-free trace of the plan. These imagined graphs are free from physical errors. Hence, comparing them with the actual scene graph helps identify errors in real robotic execution. 

\subsection{Learning the Scene-Graph Discriminator $K_\phi$}\label{subsec:sgd}
The scene graph discriminator evaluates the similarity between two scene graphs, predicting either 0 (for dissimilarity) or 1 (for similarity).  Since the edges in the scene graph are a function of the nodes, we can discriminate between two graphs by discriminating the two \textit{node} sequences.  Our graph is complete, and the objects are uniquely identifiable; thus, the pairwise comparisons are expressive enough to discriminate between the scene graphs.

Figure \ref{fig:sgp_new} illustrates the discriminator's architecture. Given two scene graphs, their corresponding pairs of nodes are concatenated and fed into an MLP to generate a score in $[0,1]$. The similarity score between the scene graphs is then calculated as the product of the individual node scores. We train the discriminator after training the $H_\psi$ and $\mathcal{T}^{ideal}_{\theta}$ using the demonstrations $\mathbb{D}$. Let $Z$ and $\widetilde{Z}$ be the scene graphs of consecutive states in a demonstration, with $o_i$ denoting the index of the moved object in the action.  
%
%
We treat pairs of the form $((o_{Z})_i, (o_{\widetilde{Z}})_j)$ as negative examples (i.e., label $= 0$), for all $i \neq j$. When $i = j$, we treat $((o_{Z})_{o_{i}}, (o_{\Tilde{Z}})_{o_{i}})$ as negative example (label $= 0$) and remaining as positive examples (label $= 1$). Training is via backpropagation of cross-entropy loss using self-supervised data from successful plan executions. 

The learnt Scene-Graph Discriminator allows failure identification during plan execution. At every action execution, the scene graph of the actual state reached, and the scene graph of the expected state predicted by the function $\mathcal{T}^{ideal}_\theta$ is compared using the discriminator. The prediction from the 
network weighed with confidence is used
to determine if the current world state deviates from the predicted, necessitating plan recovery, which we address in
the next section. Moreover, breaking down graph discrimination into nodes offers local information about the errors. The \emph{discrepancy} between two states is obtained as $\mathcal{D}: \mathcal{S} \times \mathcal{S} \rightarrow \mathbb{Z}_{\geq 0}$\textit{.} by aggregating object discrepancies as:    
$\mathcal{D}(S_1, S_2) =  n - \sum_{i = 1}^{n} K_\phi ((o_{S_1})_i, (o_{S_2})_i) 
$
Here, $o_{S_1}$ and $o_{S_2}$ are the corresponding node embeddings for $S_1$ and $S_2$, respectively, and $K_\phi$ is the learned function predicting 0 or 1 based on the similarity of the two node embeddings.
Further, this function also serves as a search heuristic while generating the recovery plan (discussed subsequently). 
%


\section{Recovery Plan Synthesis}
Given the detection of an erroneous state, we now consider the synthesis of a recovery plan. The recovery is done through a neuro-symbolic search: a symbolic search over the space of neural scene-graphs. Formally, our goal is to determine a \emph{recovery} plan, $\Pi_{S_E}$ from the erroneous state $S_E$ that reaches a nominal state on the originally intended plan $\Pi$, while minimizing the total distance to the original goal subject to a time constraint. 

Denote by $\mathtt{P}(S_k)$, the space of all plans $\Pi_K$ that reach $S_k$ from $S_E$, i.e., $\mathcal{E}(S_E, \Pi_K) = S_k$. 
Then the objective is: 
\begin{equation}
\begin{aligned}
     S_k^*, \Pi_K^*  = \argmin_{\substack{S_k \in tr(\Pi), \\ \Pi_K \in \mathtt{P}(S_k)}} \Big[\mathcal{C}_{\Pi_K}(S_{E}, S_{k}) +  \mathcal{C}_{\Pi}(S_{k}, S_{G})\Big],  
\end{aligned}
\end{equation}
subject to a planning time budget. Here, $\mathcal{C}_{\pi}(S, S^{'})$ is cost incurred in following the plan $\pi$ to reach $S^{'}$ from $S$.
The required recovery plan $\Pi_{S_E}$ is then the set of actions in $\Pi_{K}^*$ together with the actions in $\Pi$ that take from $S_k^*$ to $S_G$.

In general, we expect our errors to be local in nature, and 
we would like to exploit much of the original plan, thereby saving on re-planning time. In order to materialize this idea, we identify a set of intermediate state(s) in the trace of the given plan (referred to as sub-goals) $\{S_{t_1}, S_{t_2}, \cdots, S_{t_k}\}$ such that if we can latch on to one of these sub-goals, say, $S_{t_l}$, then we can simply follow rest of the plan from $S_{t_l}$ to $S_G$. We first describe how to efficiently re-plan from $S_E$ to given a set of sub-goals, followed by two different strategies for identifying these sub-goals in the first place.

Given a set of sub-goals $\{S_{t_1}, S_{t_2}, \cdots, S_{t_k}\}$ to latch on, we perform a multi-goal forward search to reach one of the possible subgoals. 
The minimum of the discrepancies between $S_E$ and the given subgoals is used as heuristics to guide the search. 
The following strategies are employed to prune the unrelated actions and reduce the branching factor:
\begin{enumerate}
    \item We construct a Directed Acyclic Graph (DAG), iteratively during plan execution, whose nodes represents the objects, and edges denote actions. Two objects are said to be linked via an action if they appear as arguments in some predicate corresponding to its post condition. 
    This graph aids in prioritizing object recovery by first moving those objects which appear earlier in the topological order specified by the DAG.
    \item We introduce the notion of free-space, learned using a transformer neural network, that allows us to directly predict a collision-free movement without doing an explicit search anytime an object has to be moved. The free-space transformer is provided with the bounding boxes of all objects in the scene, and the ID of the object to be moved, and it outputs a collision-free bounding box. The network is trained in a self-supervised manner where the loss consists of two components (a) MSE between the predicted free space and the bounding box of the object to ensure minimal movement (b) IoU with all other objects ensuring collision-free movement.
    \item Finally, we consider only those actions whose arguments are objects that need to be moved as identified by the discrepancy function, further pruning the search space significantly.
\end{enumerate}
%
%

We employ two strategies for determining the set of sub-goals  to plan for. First, we simply identify the state $S_t$ moving to which the error occurred and simply plan from $S_E$ to reach this singleton state. While simplistic, and effective in case of simple errors, this will fail in case of more complicated, especially cooperative errors. To address this, we devise the following strategy: we identify the set of sub-goals $\{S_{t_1}, S_{t_2}, \cdots, S_{t_k}\}$, as a set of those states whose distance to $S_E$ is minimum based on the discrepancy based heuristic. In general, the value of $k$ can be chosen based on the time available for re-planning. We can start with a small value of $k$ and keep increasing it until the planning budget is exhausted. The latter strategy is an \emph{AnyTime} recovery mechanism since (1) we increase the value of $k$ iteratively depending on the remaining planning budget (2) we stop sub-goal discovery once the planning budget is exhausted. 
\section{Evaluation Setup}\label{sec:experiments}

%
A simulated Franka Emika Panda robot~\cite{haddadin2022franka} in the PyBullet physics simulator is used to collect data set for evaluation.
Scenes are synthesized by simulating $3-5$ blocks in the physics engine by randomly sampling sizes, types and metric locations. 
The neural state transition model and the scene-graph discriminator are trained from $~3000$ atomic state transitions extracted from goal-reaching demonstrations.  
A data set of $2000+$ erroneous plan executions (see Table \ref{tab:dset-desc}) is created with the following types of execution errors: (i) grasping failures (e.g., object slipping from the gripper during picking or movement), (ii) erroneous action outcome (e.g., inexact placement of an object on top of another leading to instability and falling of a tower), (iii) motion planning errors (e.g., unexpected collision between the arm and partially constructed assembly due to uncertainty in exact object locations), (iv) actions of an external agent (e.g., a human unexpectedly moves blocks randomly or aids plan progress by pro-actively performing the next action) and (v) explicit errors (e.g., random disturbances or swapping of objects).
The set of resulting plans possess $1-10$ steps, the scenes have $5-10$ objects with a maximum of $5$ errors introduced at a single step. 

\begin{table}[ht]
    \centering
    \caption{\textbf{Evaluation Dataset Characteristics}}
    \begin{tabular}{|l|c|c|c|c|}
    \hline
     \textbf{Dataset} & \textbf{Size} & \textbf{Error Types} & \textbf{\#Objects} & \textbf{Plan len} \\
    \hline
    Data set I & 750+ & (1-5) on random step & 5 & 5 \\
    \hline
    Data set II & 1300+ & 1 on each step & 5 & 1-8 \\
    \hline
    Data set III & 200+ & 5 on random step & (6-10) & 5 \\
    \hline
    \end{tabular}
    \label{tab:dset-desc}
\end{table}

The proposed approach (\textbf{Ours}), is compared with the following baseline approaches. For a fair comparison, the action pre-conditions and association between object bounding boxes (between different states)  are assumed to be known. 
\begin{itemize}
\item \textbf{RePlan:} The PDSketch \cite{mao2022pdsketch} approach that uses a learned transition/goal-check function and performs symbolic forward search to the goal each time an error occurs. For a fair comparison the discrepancy predictor and domain specific heuristic are provided. 
\item \textbf{NoFree:} Our planner with the \textit{discrepancy}-detection function but \emph{without} the \textit{free}-space transformer network. 
\item \textbf{RL-based:} A Soft Actor Critic (SAC) agent was used to learn a goal-reaching policy in the given domain. Note that error recovery in the RL is inherent in the policy execution,  given extensive exploration during training.  
\end{itemize}
Finally, we also analyse the plan lengths arising from the anytime variant. 
Table \ref{tab:metric_table} lists the evaluation metrics, which include goal-reach-ability, length of the recovery plans and time efficiency of recovery plan synthesis.   

\begin{table}[h]
    \centering
        \caption{\textbf{Efficacy and Efficiency Metrics}}
    \begin{tabular}{|m{0.3cm}|P{2.2cm}|m{4.7cm}|}
    \hline
        & \textbf{Metric} & \parbox[c]{4.7cm}{\centering \textbf{Description}} \\
       \hline
      \multirow{3}{*}{\rotatebox[origin=c]{90}{\parbox[c]{2cm}{\centering \textbf{Correctness}}}}
       &  Detection accuracy ($Dec. \%$) &  Accuracy in detecting errors using the learnt predictor and discriminator. \\
        \cline{2-3}
        & Recovery accuracy ($Rec.\%$) & Accuracy in generating correct recovery plan given the correct error detection. \\
        \cline{2-3}
        & Goal completion rate ($Comp. \%$) & Rate of successful goal-reaching executions despite errors.   \\
        \hline
        \multirow{2}{*}{\rotatebox[origin=c]{90}{\parbox[c]{1.8cm}{\centering \textbf{Efficiency}}}} &
        Recovery plan length ($L$) & Length of the generated recovery plan from the error state to a given subgoal in the original plan. Usually measured relative to the number of errors introduced ($L/N_{err}$) and length of the most optimal recovery plan possible ($L/L_{opt}$). \\
        \cline{2-3}
        & Recovery planning Time ($T$) & Time to generate the recovery plan. Usually measured relative to the number of errors introduced ($T/N_{err}$) and length of the optimal plan ($T/L_{opt}$) \\
        
        \hline
    \end{tabular}

    \label{tab:metric_table}
\end{table}

%

\section{Results}\label{sec:results}
Our experiments evaluate (i) the effectiveness of the model in error recovery in relation to alternative approaches, (ii) an analysis of model components and (iii) its effectiveness in relation to an increase in plan length and compounding of errors, and recovering from multiple errors.

\subsection{Evaluation of Generated Recovery Plans}

\begin{figure}[ht]
    \begin{subfigure}{\hsize}
       \centering    \includegraphics[width=0.7\textwidth, height = 3.8cm]{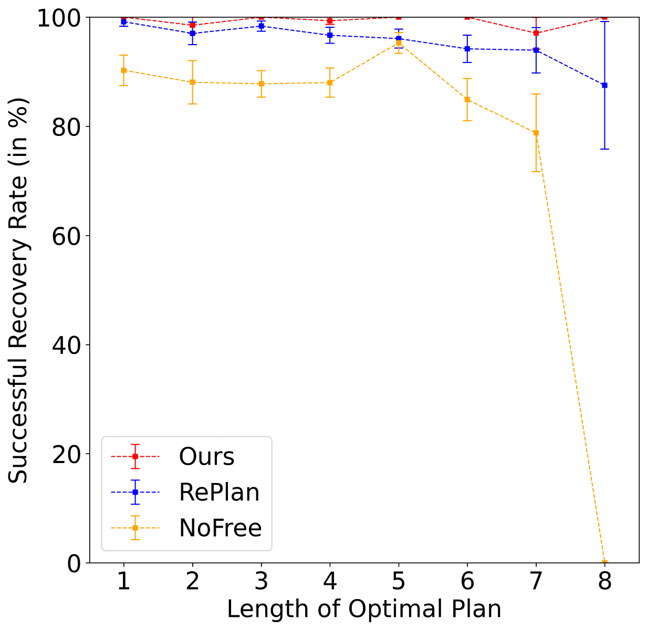}
    \end{subfigure}
    
    \begin{subfigure}{\hsize}
       \centering    \includegraphics[width=0.7\textwidth, height = 3.8cm]{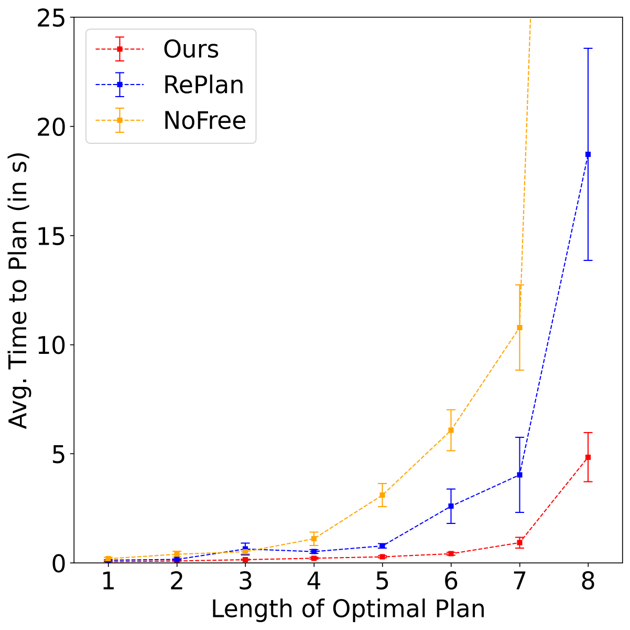}
    \end{subfigure}
    \caption{
            {\textbf{Recovery Plan Success with Increasing Errors.} Experiments analyse the recovery accuracy and time to generate a recovery plan with an increasing number of introduced errors. Note that the length of the optimal recovery plan measures the complexity of the errors.}
        }
    \label{fig:graph_increase_error}
\end{figure}

We first consider the setting where \emph{multiple errors} are introduced at a \emph{single} step during plan execution. 
Table \ref{tab:dset1} evaluates performance on \emph{data set I}, where $1$ to $5$ errors occur at a randomly chosen step. Note that recovery accuracy is reported for cases where error detection is correct, and other metrics involving $L$ and $T$ are reported when the recovery plan is correct for all the models. Figure \ref{fig:graph_increase_error} evaluates the recovery accuracy and planning time against an increasing number of errors. The length of the optimal recovery plan reflects the impact of introduced errors, with longer plans indicating more complex errors.  Our model is more effective and generates a recovery plan significantly quicker compared to the other two models, due to its discrepancy-awareness and effective action pruning from the free-space network. A low recovery time leads to a high recovery accuracy, due to the availability of an overall planning time budget ($\sim30s$). 

\begin{table}[h]
    \centering
    \caption{\textbf{Performance Comparison with Multiple Errors at a Single Execution Step}}
    \begin{tabular}{|l|c|c|c|c|c|c|}
    \hline
         \textbf{Model} & \textbf{Rec \%} &  \boldmath{$L/N_{err}$} & \boldmath{$L/L_{opt}$ }& \boldmath{}{$T/N_{err}$} & \boldmath{$T/L_{opt}$} \\ 
         \hline
         \hline
         Ours & \textbf{99.61} & 1.48 & 1.06 & \textbf{0.06s} & \textbf{0.04s} \\ 
         \hline 
         RePlan & 98.83 & 1.48 & 1.07 & 0.18s & 0.11s \\
        \hline 
         NoFree & 87.83 & 1.46 & 1.05 & 0.75s & 0.42s \\ 
         \hline
    \end{tabular}
    \label{tab:dset1}
\end{table}

\begin{figure}[ht]
    \begin{subfigure}{\hsize}
       \centering    \includegraphics[width=0.7\textwidth, height = 3.8cm]{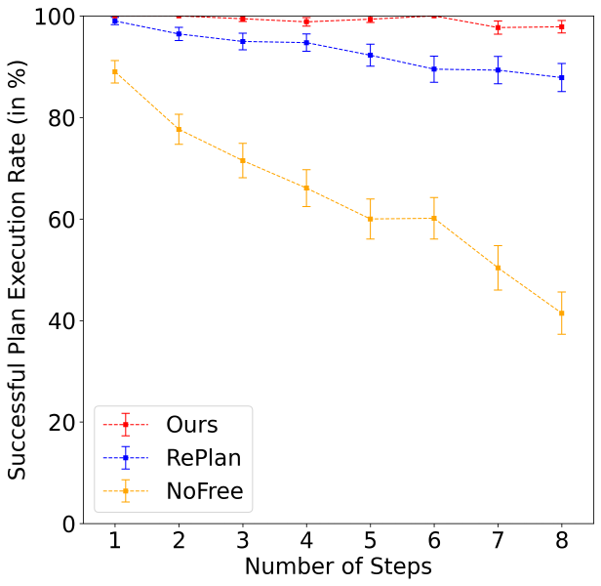}
    \end{subfigure}
    \begin{subfigure}{\hsize}
       \centering    \includegraphics[width=0.7\textwidth, height = 3.8cm]{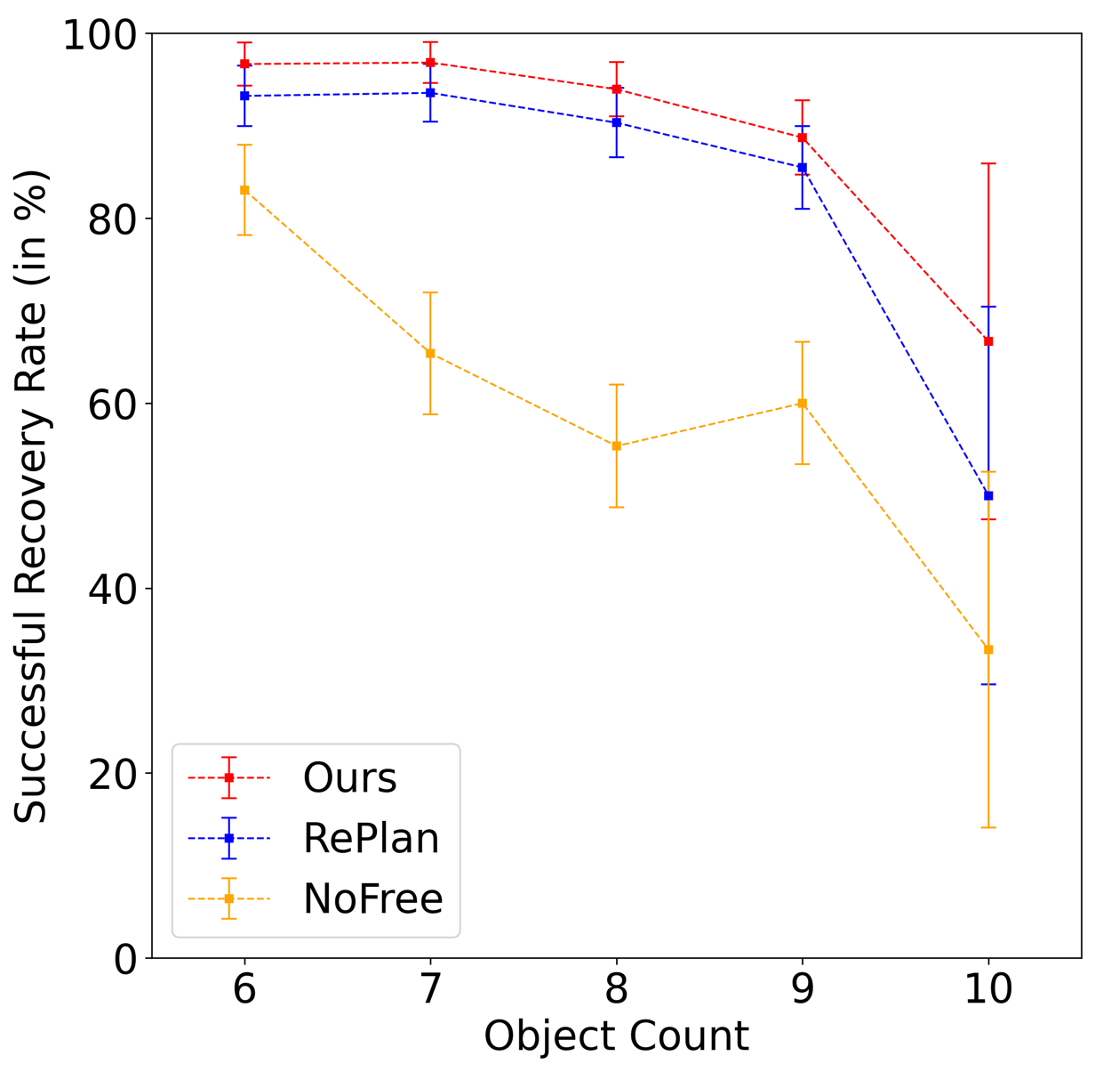}
    \end{subfigure}
    \caption{
            {\textbf{Scalability of Recovery Plan Generation Approach.} Experiments analyse the reachability of the final goal in long-horizon plans (top) and with an increasing number of objects (bottom). The proposed recovery method (in red) performs effectively for long-horizon plans and can scale to larger scenes as compared to the baselines.}
    }
    \label{fig:graph_inc_steps_objs}
\end{figure}

%
%
Next, we consider the setting where \emph{multiple errors} can occur at \emph{multiple steps} during plan execution. Table \ref{tab:dset2} reports the performance of various models on \emph{data set II}, where a single randomized error occurs after each step in the original plan ($N_{err}=1$). Figure \ref{fig:graph_inc_steps_objs} (top) shows the recovery accuracy against an increasing number of steps. Our model can recover a significantly larger part of the original plan, due to its high recovery accuracy and rapid re-planning ability. 

\begin{table}[ht]
    \centering
    \caption{\textbf{Performance with Compounding Errors}}
    \begin{tabular}{|P{0.7cm}|P{0.5cm}|P{1cm}|P{0.9cm}|P{1cm}|P{0.9cm}|P{0.6cm}|}
    \hline
         \textbf{Model} & \textbf{Rec. \%} & \boldmath$L/N_{err}$ & \boldmath$L/L_{opt}$ & 
         \boldmath$T/N_{err}$ &
         \boldmath$T/L_{opt}$ & \textbf{Comp. \%} \\ 
         \hline
         \hline
         Ours & \textbf{99.85} & 2.43 & 1.06 & \textbf{0.11s} & \textbf{0.04s} & \textbf{99.68} \\ 
         \hline 
         RePlan & 98.46 & 2.42 & 1.06 & 0.46s & 0.17s & 96.67 \\
        \hline 
         NoFree & 89.50 & 2.40 & 1.04 & 2.43s & 0.59s & 77.03 \\ 
         \hline
    \end{tabular}
    \label{tab:dset2}
\end{table}

Finally, we evaluate the generalization performance of the proposed error recovery model in environments with an increasing number of objects. A large number of objects increase the computation required for both error detection and recovery plan synthesis stages. 
Table \ref{tab:dset4} reports the performance on \emph{data set III}, where the scenes contain $6-10$ objects. We also study the variation of accuracy with an increasing number of objects in figure \ref{fig:graph_inc_steps_objs} (bottom). 
The proposed approach generalizes well to larger scenes owing primarily to discrepancy awareness and the use of learned heuristics (e.g., free space sampler) during plan search. 

\begin{table}[ht]
    \centering
    \caption{\textbf{Performance Comparison on Larger Scenes}}
    \begin{tabular}{|l|P{0.9cm}|c|c|c|c|c|}
    \hline
         \textbf{Model} & \textbf{Rec} \% &  \boldmath$L/N_{err}$ &  \boldmath$L/L_{opt}$& \boldmath$T/N_{err}$& \boldmath$T/L_{opt}$  \\ 
         \hline
         \hline
         Ours & \textbf{93.77} & 4.60 & 1.06 & \textbf{0.94} & \textbf{0.19}\\ 
         \hline 
         RePlan & 92.09 & 4.59 & 1.03 & 1.63 & 0.29 \\
        \hline 
         NoFree & 65.35 & 4.50 & 1.02 & 7.02 & 1.12 \\ 
         \hline
    \end{tabular}
    \label{tab:dset4}
\end{table}

\begin{table}[h]
    \centering
    \caption{\textbf{Recovery Plan Synthesis with Any-time Search}. The use of heuristic-guided forward search to the nearest state in the anytime version results in an improvement in re-plan quality (shorter plans). The recovery plan quality improves with increasing the re-planning budget ($K$).} 
    \begin{tabular}{|c|c|c|c|}
    \hline
         \textbf{Model} & \textbf{Rec} \% & \boldmath$L_{plan} / L_{orig}$ & \boldmath$T/N_{err}$\\ 
         \hline
         Anytime, K = 1 & 95.76 & 0.77 & 0.10s \\
        \hline 
         Anytime, K = 3 & 95.07 & 0.70 & 0.11s\\ 
         \hline
         Anytime, K = 5 & 94.52 & 0.64 & 0.15s\\ 
         \hline
    \end{tabular}
    \label{tab:dset3}
\end{table}

\begin{figure*}[h!]
    \centering

    \begin{subfigure}{\hsize}
    \centering
    \includegraphics[width=\textwidth, height = 2.8cm]{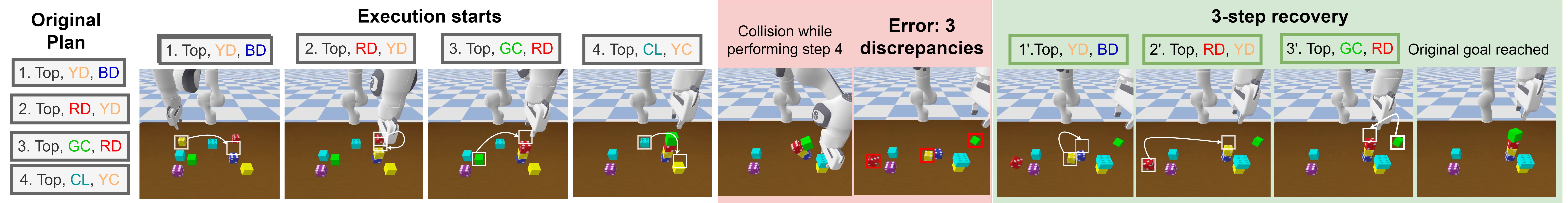}
    \caption{
    The robot follows a 4-step plan, during which a collision leads to the falling of the formed block tower. A 3-step recovery plan is generated.}
    \label{fig:qual-1}
    
    \end{subfigure}

    \begin{subfigure}{\hsize}
    \centering
    \includegraphics[width=\textwidth]{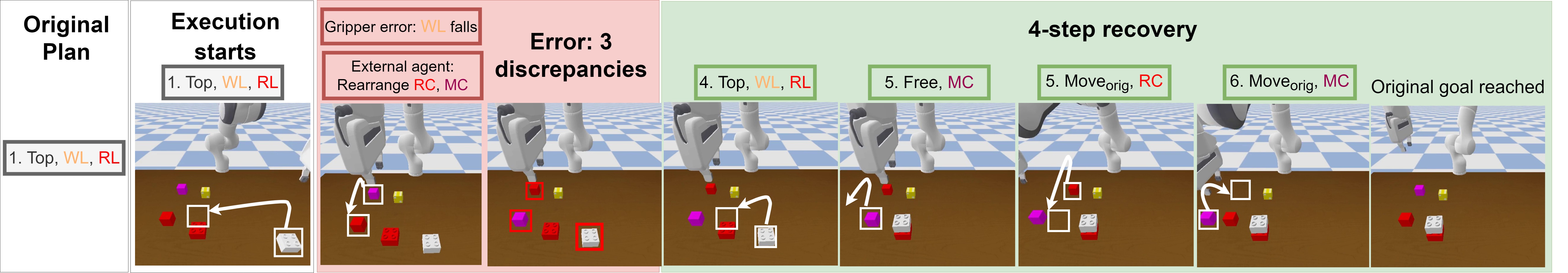}
    \caption{
    Error in robot gripper during object movement, followed by an external agent rearranging two objects, leads to an erroneous state. The 4-step recovery includes moving an object to a collision-free space.}
    \label{fig:qual-2}
    
    \end{subfigure}
    \begin{subfigure}{\hsize}
    \centering
    \includegraphics[width=0.85\textwidth, height = 3cm]{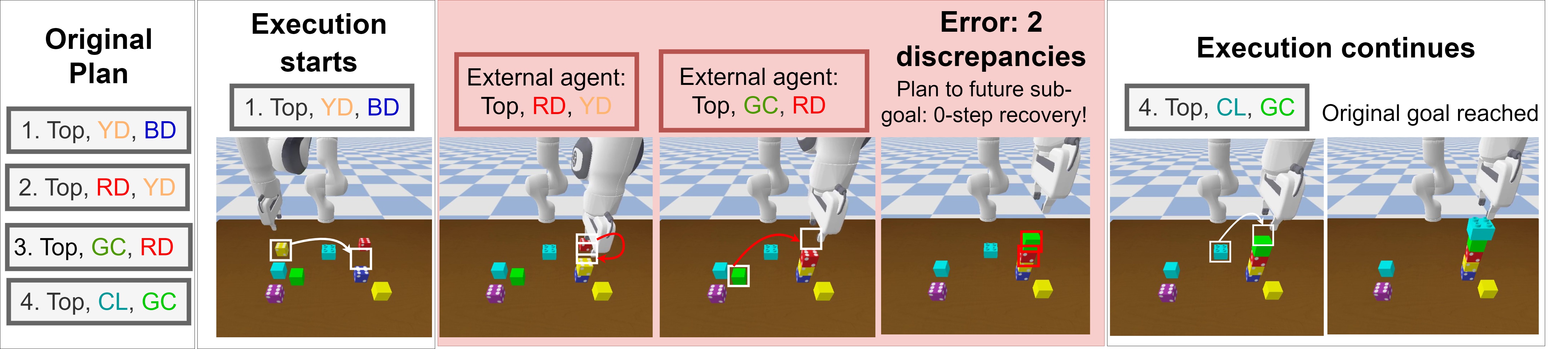}
    \caption{
    Anytime recovery: An external cooperative agent helps the robot during a 4-step plan by performing two intermediate steps. The robot identifies the constructive change and simply performs the last step of the plan.}
    \label{fig:anytime_qual}
    \end{subfigure}

    \caption{
    \textbf{Demonstration with a Franka Emika Robot in PyBullet Physics engine.} Detection of errors during plan execution and generation of recovery plans, ultimately attaining the intended goal. Errors include (a) unexpected collisions between objects and the robot, (b) grasping failure and dropping of the block during arm movement and (c) unanticipated actions by a (cooperative) human aiding one step in plan progress. 
    \emph{Abbreviations: Y=Yellow, R=Red, G=Green, B=Blue, C=Cyan, W=White, M=Magenta, D=Dice, C=Cube, L=Lego}.}
    \label{fig:qualitative_result}
\end{figure*}
\subsection{Analysis of Anytime Variant} 
For evaluating the anytime version of our approach, we consider a subset of Data set 3 containing $6-8$ objects. Table \ref{tab:dset3} reports the performance of our original and anytime algorithms with increasing value of the planning budget, $K$ as ${1,3,5}$. 
Note: in these experiments reported, the heuristic distance from a candidate sub-goal on the original plan and the true goal are approximated as being equal.
The evaluation metric $L_{plan}/L_{orig}$ represents the ratio of the length of the recovery plan (for the given anytime version of the algorithm) to the length of the recovery plan in the case of the original algorithm. 
With increasing $K$, this ratio \emph{decreases}, indicating more efficient plans. However, the time to plan \emph{increases}, contributing to a smaller recovery accuracy illustrating the inherent length-time trade-off. 

\subsection{Qualitative Results}
\begin{figure}[h]

       \centering \includegraphics[width=0.4\textwidth, height = 3cm]{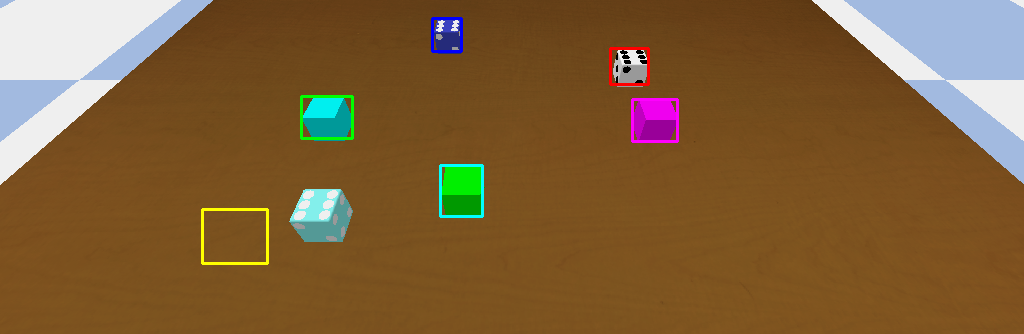}
    \caption{
        \textbf{Predictions of \emph{free}-space pose prediction module}. The task is to move the \textit{cyan lego} to a collision-free position. The learnt module predicts collision free locations (yellow bounding box) from prior data. The learning accelerates the plan search by ameliorating the need for exhaustive enumeration of possible locations.
    }
    \label{fig:free_space}
    \vspace{-0.2in}
\end{figure}

Figure \ref{fig:qualitative_result} demonstrates our approach on a simulated 7-DOF Franka Emika Panda Robot Manipulator in a simulated table-top setting. In Fig \ref{fig:qual-1}, the execution of the original 4-step plan is erroneous due to the collision of the robot arm with the tower of blocks during the last step. The discrepancy between the actual and predicted scene graphs is detected, and a 3-step recovery plan restores the blocks to their original positions.
Figure \ref{fig:qual-2} illustrates the error recovery during an adversarial scenario. During the first step of the execution, the white Lego blocks fall due to a gripper error. Moreover, an adversarial agent swaps the positions of the red and magenta cubes. Discrepancies are identified, leading to the generation of a four-step recovery plan. It's noteworthy that the magenta block must first be moved to a free space (action $[Free, MC]$) before returning the red cube to its original position (action $[Move_{orig}, RC]$), as guided by the scene-graph predictor. Additionally, figure \ref{fig:free_space} illustrates the prediction of free space in a different scene.
Finally, figure \ref{fig:anytime_qual} demonstrates the advantage of anytime recovery. After the first step of a 4-step plan, an external agent performs the next two steps. Instead of recovering to the current predicted state and unnecessary unstacking, the anytime recovery method identifies the state as being similar to the sub-goal after performing three steps, and thus, only the final step is performed.

\subsection{Accuracy of Error Detection}
%
The total error detection accuracy is $94.5\%$ ($99\%$ recall and $79\%$ precision) on the evaluation data set. The accuracies of contributing modules are: (i) $88.2\%$ for object matching, (ii) $94.2\%$ for correctly detecting the pre-conditions and (iii) $98\%$ for correct outputs of the scene graph discriminator. 
\subsection{Comparison with RL baselines}
We also evaluated an RL-based reactive planner ~\cite{li2020towards}, trained using \emph{Soft Actor Critic (SAC)} with replay buffer. The learned goal-conditioned policies performed poorly (after $12$ hours of training) in terms of goal-reaching rate, attributed largely to the domain complexity. The RL policy learnt had a 6\% success rate in predicting single-step plans for scenes containing three objects, as compared to 3.7\% by a random planner. When the scene complexity increased to five objects, the recovery rate dropped to 1.45\%. In plans with two steps, the planner could not achieve complete recovery, and demonstrated a \emph{partial} recovery rate of 1.88\%. For long-horizon plans involving more than two steps, the recovery accuracy remained zero.

\section{Conclusion}\label{sec:conclusion}
We present a discrepancy-aware neuro-symbolic approach for plan recovery from failures. Unlike existing approaches, we do not require hand-annotated data of failures, rather we make use of self-supervision to train our recovery model. 
Our approach makes use of object-centric representation of the state in the form a dense scene-graph. We train neural modules to learn the transition function based on data gathered from an existing neuro-symbolic planner. Additionally, we train neural discriminators, trained via the help other states encountered during execution as negatives, to help us distinguish the representations of the simulated state (desired) from the failure states. Once a failure is detected, a recovery plan is constructed to join back the originally constructed plan at an appropriate point. 
Incorporating exploratory actions in planning, recognizing and communicating to a human when recovery plans do not exist and  evaluation on a real robotic test bed remain part of future work.
\printbibliography

@inproceedings{bercher2014plan,
  title={Plan, repair, execute, explain—how planning helps to assemble your home theater},
  author={Bercher, Pascal and Biundo, Susanne and Geier, Thomas and Hoernle, Thilo and Nothdurft, Florian and Richter, Felix and Schattenberg, Bernd},
  booktitle={Proceedings of the International Conference on Automated Planning and Scheduling},
  volume={24},
  pages={386--394},
  year={2014}
}

@inproceedings{saetti2022optimising,
  title={Optimising the Stability in Plan Repair via Compilation},
  author={Saetti, Alessandro and Scala, Enrico},
  booktitle={Proceedings of the International Conference on Automated Planning and Scheduling},
  volume={32},
  pages={316--320},
  year={2022}
}

@inproceedings{fox2006plan,
  title={Plan Stability: Replanning versus Plan Repair.},
  author={Fox, Maria and Gerevini, Alfonso and Long, Derek and Serina, Ivan},
  booktitle={ICAPs},
  volume={6},
  pages={212--221},
  year={2006}
}

@inproceedings{rana2023residual,
  title={Residual skill policies: Learning an adaptable skill-based action space for reinforcement learning for robotics},
  author={Rana, Krishan and Xu, Ming and Tidd, Brendan and Milford, Michael and S{\"u}nderhauf, Niko},
  booktitle={Conference on Robot Learning},
  pages={2095--2104},
  year={2023},
  organization={PMLR}
}

@inproceedings{kumar2023graph,
  title={Graph inverse reinforcement learning from diverse videos},
  author={Kumar, Sateesh and Zamora, Jonathan and Hansen, Nicklas and Jangir, Rishabh and Wang, Xiaolong},
  booktitle={Conference on Robot Learning},
  pages={55--66},
  year={2023},
  organization={PMLR}
}

@article{ebert2018visual,
  title={Visual foresight: Model-based deep reinforcement learning for vision-based robotic control},
  author={Ebert, Frederik and Finn, Chelsea and Dasari, Sudeep and Xie, Annie and Lee, Alex and Levine, Sergey},
  journal={arXiv preprint arXiv:1812.00568},
  year={2018}
}

@inproceedings{li2020towards,
  title={Towards Practical Multi-object Manipulation using Relational Reinforcement Learning},
  author={Li, Richard and Jabri, Allan and Darrell, Trevor and Agrawal, Pulkit},
  booktitle={2020 IEEE International Conference on Robotics and Automation (ICRA)},
  pages={4051--4058},
  year={2020},
  organization={IEEE}
}

@article{vatsefficient,
  title={Efficient Recovery Learning using Model Predictive Meta-Reasoning},
  author={Vats, Shivam and Likhachev, Maxim and Kroemer, Oliver}
}

@inproceedings{ryu2022confidence,
  title={Confidence-Based Robot Navigation Under Sensor Occlusion with Deep Reinforcement Learning},
  author={Ryu, Hyeongyeol and Yoon, Minsung and Park, Daehyung and Yoon, Sung-Eui},
  booktitle={2022 International Conference on Robotics and Automation (ICRA)},
  pages={8231--8237},
  year={2022},
  organization={IEEE}
}

@article{thananjeyan2021recovery,
  title={Recovery rl: Safe reinforcement learning with learned recovery zones},
  author={Thananjeyan, Brijen and Balakrishna, Ashwin and Nair, Suraj and Luo, Michael and Srinivasan, Krishnan and Hwang, Minho and Gonzalez, Joseph E and Ibarz, Julian and Finn, Chelsea and Goldberg, Ken},
  journal={IEEE Robotics and Automation Letters},
  volume={6},
  number={3},
  pages={4915--4922},
  year={2021},
  publisher={IEEE}
}

@inproceedings{sung2023learning,
  title={Learning to Correct Mistakes: Backjumping in Long-Horizon Task and Motion Planning},
  author={Sung, Yoonchang and Wang, Zizhao and Stone, Peter},
  booktitle={Conference on Robot Learning},
  pages={2115--2124},
  year={2023},
  organization={PMLR}
}

@inproceedings{Kalithasan2022LearningNP,
   title     = {Learning Neuro-symbolic Programs for Language Guided Robot Manipulation},
      author    = {Kalithasan, Namasivayam and Singh, Himanshu and Bindal, Vishal and Tuli, Arnav and Agrawal, Vishwajeet and Jain, Rahul and Singla, Parag and Paul, Rohan},
      booktitle = {Proc. of ICRA},
      year      = {2023}
}

@article{Mao2019TheNC,
  title={The Neuro-Symbolic Concept Learner: Interpreting Scenes Words and Sentences from Natural Supervision},
  author={Jiayuan Mao and Chuang Gan and Pushmeet Kohli and Joshua B. Tenenbaum and Jiajun Wu},
  journal={ArXiv},
  year={2019},
  volume={abs/1904.12584}
}

@article{mao2022pdsketch,
  title={PDSketch: Integrated Domain Programming, Learning, and Planning},
  author={Mao, Jiayuan and Lozano-P{\'e}rez, Tom{\'a}s and Tenenbaum, Josh and Kaelbling, Leslie},
  journal={Advances in Neural Information Processing Systems},
  pages={36972--36984},
  year={2022}
}

@inproceedings{zhu2021hierarchical,
  title={Hierarchical planning for long-horizon manipulation with geometric and symbolic scene graphs},
  author={Zhu, Yifeng and Tremblay, Jonathan and Birchfield, Stan and Zhu, Yuke},
  booktitle={Proc. of ICRA},  
  year={2021},
  organization={IEEE}
}

@inproceedings{wang2023progport,
  author       = {Renhao Wang and
                  Jiayuan Mao and
                  Joy Hsu and
                  Hang Zhao and
                  Jiajun Wu and
                  Yang Gao},
  title        = {Programmatically Grounded, Compositionally Generalizable Robotic Manipulation},
  booktitle    = {The Eleventh International Conference on Learning Representations,
                  {ICLR}},
  year         = {2023},
  url          = {https://openreview.net/pdf?id=rZ-wylY5VI},
}

@INPROCEEDINGS{He2016Resnet,
  author={He, Kaiming and Zhang, Xiangyu and Ren, Shaoqing and Sun, Jian},
  booktitle={2016 IEEE Conference on Computer Vision and Pattern Recognition (CVPR)}, 
  title={Deep Residual Learning for Image Recognition}, 
  year={2016},
  volume={},
  number={},
  pages={770-778},
  doi={10.1109/CVPR.2016.90}}

@inproceedings{huang2019neural,
  title={Neural task graphs: Generalizing to unseen tasks from a single video demonstration},
  author={Huang, De-An and Nair, Suraj and Xu, Danfei and Zhu, Yuke and Garg, Animesh and Fei-Fei, Li and Savarese, Silvio and Niebles, Juan Carlos},
  booktitle={Proceedings of the IEEE/CVF conference on computer vision and pattern recognition},
  pages={8565--8574},
  year={2019}
}

@inproceedings{shridhar2022cliport,
  title={Cliport: What and where pathways for robotic manipulation},
  author={Shridhar, Mohit and Manuelli, Lucas and Fox, Dieter},
  booktitle={Conference on Robot Learning},
  pages={894--906},
  year={2022},
  organization={PMLR}
}

@inproceedings{li2021reactive,
  title={Reactive task and motion planning under temporal logic specifications},
  author={Li, Shen and Park, Daehyung and Sung, Yoonchang and Shah, Julie A and Roy, Nicholas},
  booktitle={2021 IEEE International Conference on Robotics and Automation (ICRA)},
  pages={12618--12624},
  year={2021},
  organization={IEEE}
}

@article{haddadin2022franka,
  title={The franka emika robot: A reference platform for robotics research and education},
  author={Haddadin, Sami and Parusel, Sven and Johannsmeier, Lars and Golz, Saskia and Gabl, Simon and Walch, Florian and Sabaghian, Mohamadreza and J{\"a}hne, Christoph and Hausperger, Lukas and Haddadin, Simon},
  journal={IEEE Robotics \& Automation Magazine},
  volume={29},
  number={2},
  pages={46--64},
  year={2022},
  publisher={IEEE}
}

@article{xu2019regression,
  title={Regression planning networks},
  author={Xu, Danfei and Mart{\'\i}n-Mart{\'\i}n, Roberto and Huang, De-An and Zhu, Yuke and Savarese, Silvio and Fei-Fei, Li F},
  journal={Advances in Neural Information Processing Systems},
  volume={32},
  year={2019}
}

@inproceedings{knepper2013ikeabot,
  title={Ikeabot: An autonomous multi-robot coordinated furniture assembly system},
  author={Knepper, Ross A and Layton, Todd and Romanishin, John and Rus, Daniela},
  booktitle={2013 IEEE International conference on robotics and automation},
  pages={855--862},
  year={2013},
  organization={IEEE}
}

@article{tellex2014asking,
  title={Asking for help using inverse semantics},
  author={Tellex, Stefanie and Knepper, Ross and Li, Adrian and Rus, Daniela and Roy, Nicholas},
  year={2014},
  journal={Robotics: Science and Systems Foundation}
}

@inproceedings{hudson2012end,
  title={End-to-end dexterous manipulation with deliberate interactive estimation},
  author={Hudson, Nicolas and Howard, Thomas and Ma, Jeremy and Jain, Abhinandan and Bajracharya, Max and Myint, Steven and Kuo, Calvin and Matthies, Larry and Backes, Paul and Hebert, Paul and others},
  booktitle={2012 IEEE International Conference on Robotics and Automation},
  pages={2371--2378},
  year={2012},
  organization={IEEE}
}
\addtolength{\textheight}{-12cm}   


\end{document}